\documentclass{article}
    \usepackage{arxiv}
    \usepackage{multirow}
    \usepackage{moreverb,url}
    \usepackage{amssymb}

\usepackage{graphicx}
\usepackage{amsmath,stmaryrd}
\usepackage{booktabs}

    \usepackage[backend = bibtex, style=authoryear]{biblatex}
    \DeclareLanguageMapping{british}{british-apa}
    \usepackage[pdftex]{hyperref}

\usepackage{calc}
\newlength{\eqhsize}
\newcommand{\myinlineeq}[1]{%
    \setlength{\eqhsize}{\minof{\widthof{\mbox{\ensuremath{#1}}}}{\linewidth}}
    \resizebox{\eqhsize}{!}{\ensuremath{#1}}}

\def\eg{e.g.,\ }

\begin{document}

\title{Temporal Signals to Images: Monitoring the Condition of Industrial Assets with Deep Learning Image Processing Algorithms}

    \author{%
    	Gabriel Rodriguez Garcia\\
        ETH Z\"urich,\\
        Z\"urich, Switzerland\\
        \And 
        Gabriel Michau\\
        ETH Z\"urich,\\
        Z\"urich, Switzerland\\
        \And
        M\'elanie Ducoffe\\
        Airbus AI Research,\\
        Toulouse France\\
        \And
        Jayant Sen Gupta\\
        Airbus AI Research,\\
        Toulouse France\\
        \And 
        Olga Fink\\
        ETH Z\"urich,\\
        Z\"urich, Switzerland}
        \subtitle{Preprint - Published version: \url{https://doi.org/10.1177/1748006X21994446}}
        \date{21st of February 2021}
       \maketitle

\begin{abstract}%
The ability to detect anomalies in time series is considered highly valuable in numerous application domains. The sequential nature of time series objects is responsible for an additional feature complexity, ultimately requiring specialized approaches in order to solve the task. 
Essential characteristics of time series, situated outside the time domain, are often difficult to capture with state-of-the-art anomaly detection methods when no transformations have been applied to the time series.
Inspired by the success of deep learning methods in computer vision, several studies have proposed transforming time series into image-like representations, used as inputs for deep learning models, and have led to very promising results in classification tasks.

In this paper, we first review the signal to image encoding approaches found in the literature. Second, we propose modifications to some of their original formulations to make them more robust to the variability in large datasets. Third, we compare them on the basis of a common unsupervised task to demonstrate how the choice of the encoding can impact the results when used in the same deep learning architecture. 
We thus provide a comparison between six encoding algorithms with and without the proposed modifications. The selected encoding methods are Gramian Angular Field, Markov Transition Field, recurrence plot, grey scale encoding, spectrogram, and scalogram. We also compare the results achieved with the raw signal used as input for another deep learning model. We demonstrate that some encodings have a competitive advantage and might be worth considering within a deep learning framework.

The comparison is performed on a dataset collected and released by Airbus SAS \cite{airbus18}, containing highly complex vibration measurements from real helicopter flight tests. 
The different encodings provide competitive results for anomaly detection.

\end{abstract}

    \keywords{Unsupervised Fault Detection; Time Series Encoding; Helicopters; Vibrations; CNN.}
    \maketitle

\section{Introduction}
\label{sec:intro}

With the globalisation of many industrial markets, it is becoming increasingly important for operators to optimise the use and operating costs of their assets. This optimisation requires the ability to maintain a high availability of the assets at regulatory safety levels. Since the maintenance of industrial assets accounts for a large part of the operating costs, the ability to monitor the system in order to perform maintenance based on its present condition would benefit the operators greatly~\parencite{firdaus2019optimization}. The ability to detect anomalies in real time could help in the short term to reduce the necessary amount of unscheduled maintenance. Once it has been demonstrated that the algorithms are able to detect anomalies reliably, these same algorithms could also help to reduce the amount of preventive maintenance required. 

With the recent decrease in the cost of sensors and the possibility to connect large monitoring environments with the industrial Internet of Things, large data streams are routinely captured and can be accessed in real time, even for high frequency data. Even today, the implementation of condition-based maintenance is often lacking the tools to automatically and reliably process such large quantities of information in real time~\parencite{li2020review}. In fact, the handling of high-frequency data is facing very specific challenges, such as the variable length of the data streams and the fact that the fault signatures can appear at completely different temporal or spectral scales, for example over several hours of continuous operation monitored at several thousand or millions of hertz~\parencite{Valle2009Streaming,al2020industrial}. Even for experts, it may be difficult to define up-front the relevant temporal or spectral features for all the possible faults and anomalies. 

Traditionally, condition monitoring data are handled with feature engineering. Yet devising relevant features is a task specific to each dataset. Finding the right features remains an open research problem in unsupervised context where it is not possible to select nor tune \textit{a priori} the features based on the anomaly contained in the dataset. To the best of our knowledge, there is no pipeline (feature generation, unsupervised feature selection, anomaly detection) that is widely considered as a reference by the community. These limitations in the generalisability of feature engineering approaches are a further encouragement to look for automated end-to-end pipelines.

The recent development of deep learning provides the possibility to automatically learn from large datasets the signatures that will maximise the network's training objective and offers state-of-the-art results in supervised setups~\parencite{saufi2019challenges,qu2019gear,fawaz2019deep}.  In the industry, however, for some complex systems, it is not possible to cover all possible types of fault modes and, therefore, similarly infeasible to collect representative samples to handle their detection in a supervised manner. While some studies try to define the problem as a supervised classification task, proposing solutions for imbalanced datasets~\parencite{Jia2018}, supervised learning approaches will still not be applicable to cases of new or not-yet-observed fault types, which is a common phenomenon in complex systems. Consequently, an alternative is to train an anomaly detector on data collected exclusively in healthy operating conditions in order to detect any deviation from the healthy state~\parencite{michau2020feature}.  Indeed, complex industrial systems operate almost exclusively in healthy conditions since they are reliable by design, preventively maintained, and shut down when conditions worsen in order to avoid major failure modes. Previous works have shown that monitoring the residuals of an auto-encoder trained on healthy condition monitoring data could efficiently be used as a means of anomaly detection~\parencite{Hu2017Fault,Chouhan2019Network}. 

Recent studies have also demonstrated that when dealing with signals, their transformation into images could provide competitive results in supervised setups~\parencite{gondara2016medical,krummenacher2017wheel,sasirekha2020novel}. In these works, the authors show that encoding time series to images can help to emphasize, capture, or condense local patterns that would otherwise be spread over time. In the literature, we identified four promising encoding algorithms: the Gramian Angular Field, the Markov Transition Field~\parencite{wang2015encoding}, the recurrence plot~\parencite{MARWAN2007237}, and grey scale encoding~\parencite{xu2019online}. These are in addition to the traditional visualisation tools used in signal processing: the spectrogram and the scalogram~\cite{boashash2015time,verstraete2017deep}. To the best of our knowledge, however, these approaches have never been tested on anomaly detection tasks for time series, despite major successes in the use of deep learning auto-encoders in image processing~\parencite{zhang2019ae2}. Also, these encodings have not yet been simultaneously evaluated on a single task.

\begin{figure*}
\centering
\includegraphics[width=\textwidth]{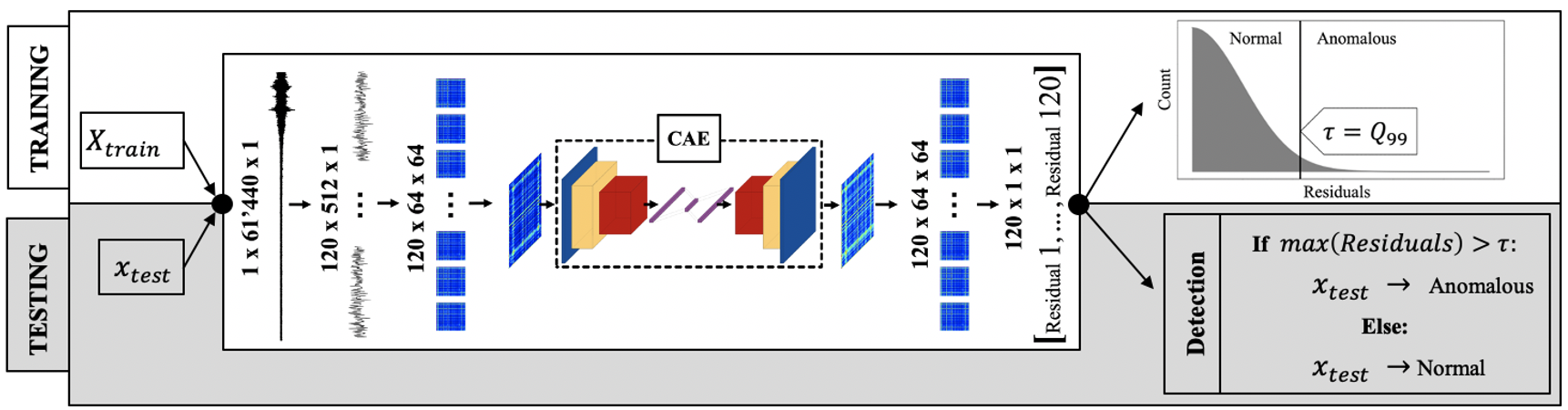}
\caption{\textbf{Framework}. The network is trained on N time series in the training dataset $X_{train}$ and the threshold is set as the 99th percentile over the sample of 120$\times$N residuals (108$\times$N for the GS encoding). For testing an arbitrary test sample $x_{test}$, each of the 120 slices (108 for the GS encoding) is reconstructed by the CAE and the largest residual is compared with the threshold $\tau$ to detect anomalies. When using the raw time series, the same framework is followed without the encoding step and with 1-D convolutions in the auto-encoder.\label{fig:framework} }
\end{figure*}

In this paper, we propose to  tackle these two gaps by comparing  these six signal to image encoding algorithms in an unsupervised setup. Since some encoding algorithms, in their original formulation, were invariant to some signal transformations, such as amplitude changes and shifts, we additionally propose alternative formulations. The framework for the comparison is described in Figure~\ref{fig:framework} and consists of the following steps: first, the time series splitting; second, the encoding of each sub-series; and last, the monitoring of their aggregated residuals when fed through a convolutional auto-encoder (CAE). We compare the results with a similar framework, taking the raw sub-series as input and using a one-dimensional temporal CNN. Temporal CNNs have proven superior to simple dense networks in signal processing with deep learning~\parencite{Sokolovsky2018deep}, while also being much lighter and simpler to use than recurrent networks~\parencite{aksan2018stcn}. Furthermore, using temporal CNN makes the network comparable to the convolutional neural networks used to process the encoded images. 
Our results demonstrate that, first, the suggested modifications to the encoding methods are required to obtain meaningful results and that, second, the encoding algorithms can offer a competitive advantage for the detection of anomalous time series. Lastly, we discuss the computational requirements of the proposed framework.

The remainder of this paper is organised as follows: Section~\ref{sec:anode} describes the proposed anomaly detection framework; Section~\ref{sec:encod} presents the different image encodings used to represent time series data; this methodology is applied with the different image encodings to an industrial use case introduced in Section~\ref{sec:casestudy}; and finally, we discuss the methods and perspectives.

\section{Framework for Anomaly Detection}
\label{sec:anode}

The proposed framework, illustrated in Figure \ref{fig:framework}, entails the detection of anomalous input reconstruction based on an auto-encoder neural network with different types of time series-to-image encodings as input. The underlying idea is to train an auto-encoder to reconstruct its own input while compressing it to a lower dimensional latent space, acting as a bottleneck. The network will learn to reconstruct as well as possible its input data (either time series or images), following the same distribution as those it has been trained on.
If the the auto-encoder is trained on healthy data, its reconstruction ability will be poor if presented with anomalous conditions that deviate from the healthy conditions used during training. Monitoring the reconstruction error with respect to a threshold allows for the detection of such anomalous patterns. 

Handling long time series raises two types of challenges. First, since the monitored residual is an aggregated measure over the reconstruction, localised anomalies may be unnoticeable at the global scale. Second, for most encodings, computational time and size evolve according to a power law of the input size. To overcome these two challenges, the time series are subdivided into smaller sub-series of size $l_1$. These sub-series are then transformed by one of the signal encoding algorithms to images of size $l_2\times l_2$, used as input for the auto-encoder.  Their residuals are computed and aggregated. We define the residual of the $k$-th sub-series, $\mathrm{Res}_k$, as the $\ell_1$-norm of the difference between the input, $x_{\mathrm{or}}$, of size $l$ and its reconstruction, $x_{\mathrm{rc}}$ as:
\begin{equation}
    \mathrm{Res}_k = \sum_{i=1}^{l}(|x_{\mathrm{or}}^i - x_{\mathrm{rc}}^i|),
    \label{eq:residual}
\end{equation}
where $l=l_1$ for the baseline model using the raw sub-series without any encodings as input to a one-dimensional CNN, and $l=l_2\times l_2$ for all other cases.

During training, only time series from healthy conditions are used. Based on the distribution of the residuals of all sub-series, we extract the 99th percentile to define the detection threshold $\tau$. In the test set, for each time series, we monitor the maximum residual over its sub-series, and compare it to the threshold $\tau$. If it exceeds the threshold, the time series is detected as anomalous. Monitoring the maximum of the residuals enables the detection of local anomalies that would impact even one of the sub-sequences only.

\section{Time Series-to-Image Encoding}
\label{sec:encod}

    \subsection{Gramian Angular Field (GAF)}

The Gramian Angular Field~\parencite{wang2015encoding} consists of a two-step process to transform a time series into a matrix. First, the transformation from the space (\textit{Time}$\times$\textit{Value}) to polar coordinates is performed. Then, the Gramian matrix in that new space is computed.

The transformation to polar coordinates is accomplished through the following operation:
\begin{equation}
\label{eq:GAF-pol}
     \begin{cases}
              \phi = \arccos(X_{norm,i}) ; \phi \in [0,\pi],
              \\
              r = \frac{n}{L}; r \in \mathbb{R}^+,
     \end{cases}
\end{equation}
where L is scaling the temporal dimension but does not influence the final matrix. This transformation is bijective and absolute temporal relations are preserved~\parencite{wang2015encoding}.
Since Equation~\ref{eq:GAF-pol} is only defined for $X\in [-1,1]$, it first requires the scaling of the data. \cite{wang2015encoding} propose scaling the input $X$ as 
    \begin{equation}
    X = (x_1,x_2,...,x_N),\quad
    X_{norm,i} = \frac{x_i - \mathrm{UB} + (x_i - \mathrm{LB})}{\mathrm{UB} - \mathrm{LB}} ; \in [-1,1],
    \end{equation}  
where UB and LB refer to the upper and lower bound parameters, defined originally as $\mathrm{UB}=\mathrm{max}(x)$ and $\mathrm{LB}=\mathrm{min}(x)$.
The final Gramian matrix can be computed as follows:
\begin{equation}
\myinlineeq{%
               GAF = {X_{norm}}^{T} \cdot X_{norm} - \sqrt{I-{X_{norm}}^2}^{T}\cdot \sqrt{I-{X_{norm}}^2}}
\end{equation}
\begin{equation}
\myinlineeq{%
                  GAF =  \begin{pmatrix}
                          \cos(\phi_1+\phi_1) & \cos(\phi_1+\phi_2) & \cdots & \cos(\phi_1+\phi_N)\\
                          \cos(\phi_2+\phi_1) & \cos(\phi_2+\phi_2) & \cdots & \cos(\phi_2+\phi_N) \\
                          \vdots  & \vdots  & \ddots & \vdots  \\
                          \cos(\phi_N+\phi_1) & \cos(\phi_N+\phi_2) & \cdots & \cos(\phi_N+\phi_N)
                         \end{pmatrix}
                         }
\end{equation}
where $I$ refers to a unit row vector and the superscript $(\cdot)^T$ refers to the transpose operator. By analogy to Gram matrices $G_{ij} = \langle x_i,x_j\rangle$, where $\langle\cdot,\cdot\rangle$ is an inner product, the matrix is called a Gramian Angular Field. One may observe, however, that $\cos(x_i+x_j)$ does not satisfy all the conditions of inner-products. In addition, it should be noted that the matrix does not depend on $r$. Therefore, it depends neither on time nor on the choice of $L$ in Eq~\ref{eq:GAF-pol}. As a consequence, the GAF transformation is not reversible, meaning that the GAF no longer allows for a unique reconstruction of the time series. This leads to some loss of information after the encoding has been applied. However, since there is often only moderate overlap in the inverse space, the input can be roughly approximated \parencite{wang2015encoding}.

\textbf{Proposed modification:} In our study, we opt to scale the training samples based on the full training dataset in order to keep the relationships and differences between the samples. Therefore, we propose to define $\mathrm{UB}$ and $\mathrm{LB}$ based on the training set distribution.
If, in the test set, a sample contains values exceeding these bounds, we propose to clip the values to the bound.

    \subsection{Markov Transition Field (MTF)}
The Markov Transisition Field \parencite{wang2015encoding} entails building the matrix of probabilities in order to observe the disparity in value between any pair of points in the time series. 
First, the time series is discretized. Then, the transitions $w$ from one bin to another are counted and normalised over every two consecutive data-points in the whole training set. Lastly, the MTF matrix contains for every pair of points in the time series the transition probability between the two bins the points belong to.

Assuming a discretization of the time series based on $Q+1$ bin edges, that is, $Q$ bins $q_i$, the transitions are computed over every two consecutive points in the training set as
\begin{equation}
  \forall (i,j)\in \llbracket 1,Q+1\rrbracket,\quad \widehat{w}_{ij} = \text{number of transitions } q_i \rightarrow q_j
\end{equation}
\begin{equation}
  \forall (i,j)\in \llbracket 1,Q+1\rrbracket,\quad w_{ij} = \frac{\widehat{w}_{ij}}{\sum_j \widehat{w}_{ij}}
\end{equation}
$W =(w_{ij})$ is the Markov Transition matrix, which at this stage does not capture any temporal relations. 
The final $N\times N$ matrix, called the Markov Transition Field, is computed as follows:
\begin{equation}
\myinlineeq{%
               MTF =   \begin{pmatrix}
                      w_{ij | X_1 \in q_i, X_1 \in q_j}  & \cdots & w_{ij | X_1 \in q_i, X_N \in q_j}\\
                      w_{ij | X_2 \in q_i, X_1 \in q_j} & \cdots & w_{ij | X_2 \in q_i, X_N \in q_j} \\
                      \vdots  & \ddots & \vdots  \\
                      w_{ij | X_N \in q_i, X_1 \in q_j} & \cdots & w_{ij | X_N \in q_i, X_N \in q_j}
                     \end{pmatrix}
                     }
\end{equation}

This transformation of the time series is non-reversible, meaning that there is substantial information loss. In particular, the discretization of the time series plays a crucial role in the amount of information kept or lost by the transformation.  A large number of bins might lead to sparsity in the image, while a small number of bins might lead to substantial information loss~\parencite{wang2015encoding}.

\textbf{Proposed modification:} 
In the original formulation of the MTF, \cite{wang2015encoding} proposed partitioning the range of values into a number of equi-sized segments. Since in many datasets, the distribution of values tends to follow long-tailed distributions, choosing the optimal number and width of bins becomes a challenging step requiring manual tuning. A large bin width would aggregate most values into the bins the closest to the mean, while a small bin width would lead to sparsity in the extreme bins.

To overcome this high dependency on the optimal parameter setting, we propose to extend the original formulation of the bin choice and assignments, using instead the Symbolic Aggregate approXimation (SAX) algorithm~\parencite{Li2007}, which performs a non-uniform bin assignment such that the distribution of the bin frequency roughly follows a Gaussian distribution. As such, setting the right number of bins is a less critical task, since the risk of sparsity is drastically reduced. An arbitrarily large number of bins can be used, and the choice mostly depends on the computational requirements for the transitions.

    \subsection{Recurrence Plot (RP)}
    
The recurrence plot (RP) \parencite{kamphorst1987recurrence} transforms a time series into a matrix of recurrences to reveal at which points some trajectories return to a previously visited state.
In this paper, we follow a non-binarized version of RP, as proposed by \cite{souza2014extracting}. The RP encoding consists of the following transformation:
\begin{equation}
\mathrm{RP}_{ij} = ||S_i-S_j||,\quad i,j\in \llbracket1,K\rrbracket,
\end{equation}
where the time series is split into $K$ sub-sequences $(S_i)_{i\in \llbracket1,K\rrbracket}$. Here, we set the sub-sequence length to $1$ (we only consider the difference between single values), and defined $K=l_2$.

\textbf{Proposed modification:} 
This transformation is non-reversible and also loses information on the values in the time series since only the differences in values are kept. We propose, therefore, the following transformation:
\begin{equation}
               \widehat{\mathrm{RP}} = \mathrm{RP} + \mathrm{mean}(X).
\end{equation}
This transformation aims at capturing recurrences in individual time series around their mean.

    \subsection{Gray-Scale (GS) Encoding}
    
Grey scale encoding has been revisisted in \cite{wen2018} for the purpose of fault diagnosis in manufacturing systems using convolutional neural networks. The transformation consists of, first, the reshaping of a 1D time series into a collection of $K$ sub-series of size $K_1$ and, second, the rescaling of the values as color encoding values (\eg 8-bit integer). In this work, for consistency with other encodings, we aim at $K=K_1$ and compute the stride $s$ between the start of each sub-series accordingly. The GS encoding corresponds, therefore, to the following operation:
\begin{equation}
              GS_{i,j} = \mathrm{round}\left(P \cdot \frac{(x_{(i-1) \cdot s + j} - \mathrm{LB})}{\mathrm{UB} - \mathrm{LB}}\right),\ \ i,j\in \llbracket1,K\rrbracket,
\label{eq:gs}
\end{equation}
where the operator $\mathrm{round}$ rounds the value to the nearest integer, $P$ is a scaling factor, and $\mathrm{UB}$ and $\mathrm{LB}$ are respectively an upper and lower bound to be used for the scaling of $X$.

\textbf{Proposed modification:} 
In the original formulation, the authors suggested using $\mathrm{UB}=\mathrm{max}(X)$, $\mathrm{LB}=\mathrm{min}(X)$ and $P=255$ such that the resulting values can be interpreted as 8-bit encoded grey scale images. In this work, we propose first defining $\mathrm{UB}$ and $\mathrm{LB}$ over the entire training dataset, choosing these bounds in accordance with the GAF encoding. Furthermore, since most common deep learning implementations use 32-bit inputs, we recommend evaluating the necessity of the conversion to integers for this encoding.

    \subsection{Spectrogram (SP) and Scalogram (SC)}
    
To complement the encodings described above, we suggest adding to the study a more traditional time-frequency analysis. Since the spectro-temporal description of a time series provides two-dimensional representations, it is customary in the literature to interpret these representations directly as images~\parencite{verstraete2017deep}. In this work, we propose two popular transformations: the spectrogram, resulting from the short-time Fourier transform (STFT), and the scalogram, resulting from the discrete wavelet transform (DWT). For more information on these transformations, the reader is referred to the book from \cite{boashash2015time}.

Both the STFT and the CWT rely on the principle of convolving the time series with a windowing function. The difference between the two approaches is that for STFT, the window is fixed and the Fourier transfrom is applied to the convolved signal, while for the CWT, the window is based on a wavelet function, whose scaling enables the exploration of the spectrum at different levels. For the STFT, it is therefore necessary to set three hyperparameters: the window type, its size, and its temporal stride. The DWT requires the a-priory definition of the wavelet family, the scales explored, and the temporal stride.

\section{Detection of Anomalous Conditions in Flight Test}
\label{sec:casestudy}
    \subsection{Helicopter accelerometers use case}
    The use case on which we illustrate different ways to encode time series into images has to do with flight test helicopters' vibration measurements. The dataset has been collected and released by Airbus SAS \cite{airbus18}\footnote{\url{https://doi.org/10.3929/ethz-b-000415151}}. A main challenge in flight tests of heavily instrumented aircraft (helicopters and airplanes alike) is the validation of the generated data because of the large number of signals to validate. Manual validation requires too much time and manpower. Therefore, automation of this validation is crucial. 
    
    In this case, accelerometers are placed at different positions on the helicopter and in different directions (longitudinal, vertical, lateral) to measure the vibration levels in all operating conditions of the helicopter. This implies different vibration levels in the dataset. The dataset consists of multiple 1D time series with a constant frequency of 1024 Hz taken from different flights, cut into 1 minute sequences. Labels (normal, anomalous) are associated per flight and attributed to each sequence of the flight.
    
    The dataset has two parts:
    \begin{itemize}
        \item \underline{the training dataset}: consists of 1677 sequences from normal flights; 
        \item \underline{the test dataset}: consists of 594 sequences from both normal and anomalous flights (297 of each).
    \end{itemize}
    
    The challenge here is to train the algorithms on the healthy condition monitoring data from the training dataset that are able to reliably detect anomalies in the test dataset (without any access to the anomalous time series during training and without any prior knowledge concerning the number and types of possible anomalies). Finding optimal network architectures and optimal hyperparameters is an unresolved research question for unsupervised setups and is not the focus of this paper. Since at training time, it is assumed that only healthy data are available, hyperparameters cannot be validated based on their ability to detect anomalies. Their choice has somehow to rely on external and thus partly arbitrary justifications such as the trade-off between information loss and computational requirements, or application-oriented justifications.
    
    \subsection{Pre-processing and Hyperparameter Settings for the Time Series-to-Image Encodings}

\begin{figure}
\centering
\includegraphics[width=8cm]{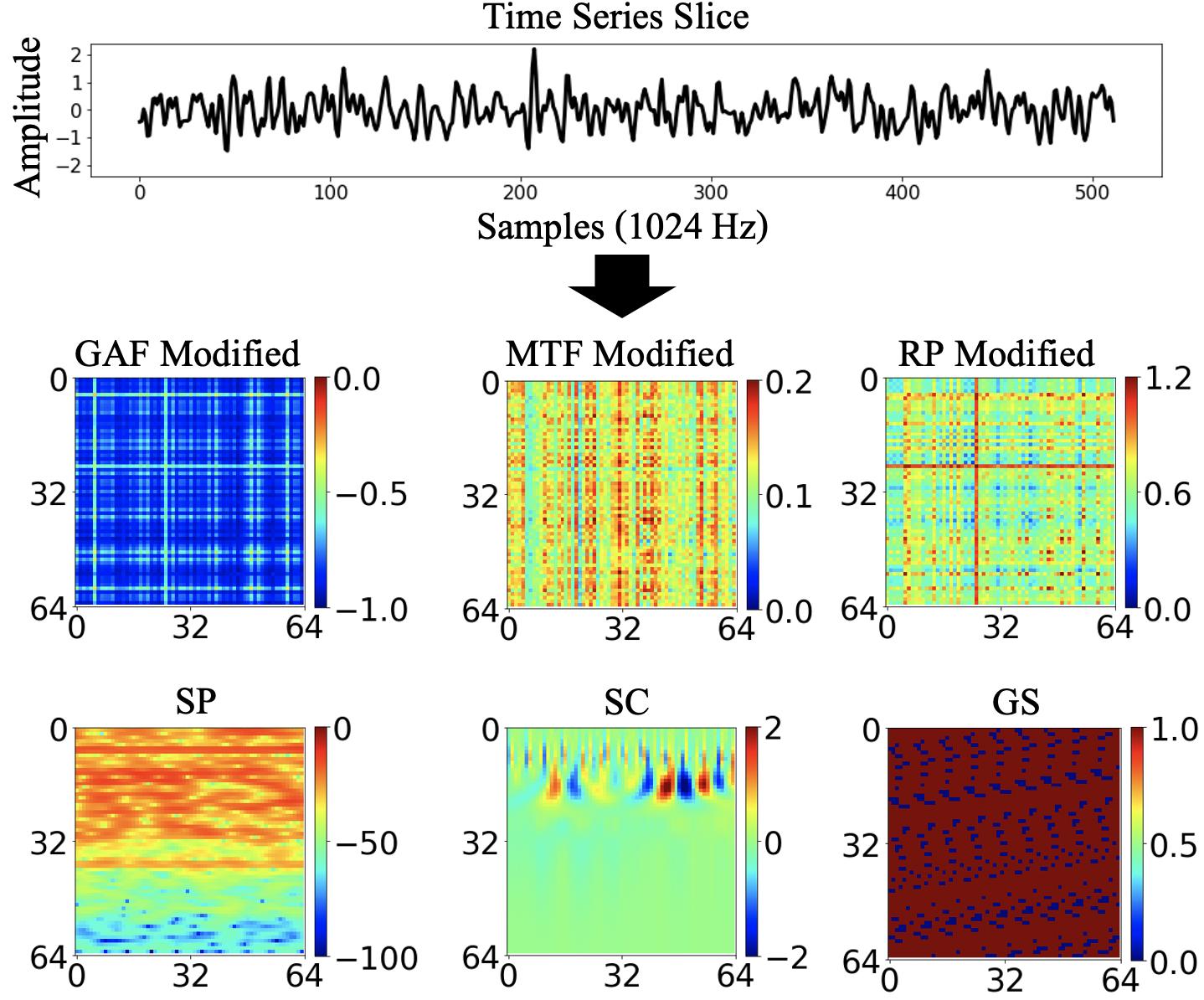}
\caption{\textbf{Encoding Examples}. The colors indicate the value of the encoding according to the corresponding colorbars. Each encoding lead to a $64\times 64\time 1$ map.} \label{fig:examples}
\end{figure}

To evaluate the benefits of using the various time series-to-image encodings, we compare the six selected encodings and their extensions to one another but also to a baseline that consists of raw 1D time series inputs without any encoding. For the 1D input, a 1D-CNN is applied instead of the 2D-CNN that is applied for all the time series-to-image encodings. As illustrated in the framework designed in Figure~\ref{fig:framework}, we split the time series of length 61\,440 into 120 slices of size $l_1$ equal to 512, such that each slice represents half a second of signal monitoring, a temporal resolution deemed relevant for the application at hand. No additional operations, such as smoothing, outlier detection, or normalisation, are performed.

For all proposed encodings, we generate one image for each of the 120 sub-series of size 512. We set all parameters, when possible, to obtain images of size 64$\times$64, as a trade-off between the level of detail contained in the images and the computing efficiency. The GAF, MTF, and RP encodings have, by definition, a resulting image size equal to that of the input time series. Therefore, we downscale the resulting images using average pooling.

In the following, the hyperparameter settings for the selected encodings are described.
    
\textbf{GAF:}
The GAF has only a single parameter to tune, which is the clipping parameter that determines the allowable upper and lower boundaries for time series values. 
Following the work in \cite{beirlant2006statistics}, we find a linear relationship in the QQ-plot of the training dataset. By extrapolation, we find that the probability of observing a value beyond $1.2$ times the max or the min of $X$ is numerically equal to $0$. 
We therefore choose these values as the upper and lower bounds $\mathrm{UB}$ and $\mathrm{LB}$. Values exceeding this threshold in the test set are clipped.

\textbf{MTF:}
\begin{figure}
\centering
\includegraphics[width=0.8\columnwidth]{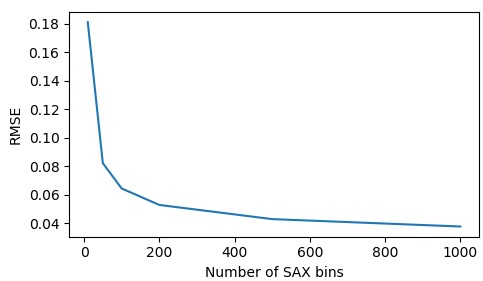}
\caption{RMSE of SAX inverse transformation versus the number of bins used in the SAX algorithm}
\label{fig:SAXbins}
\end{figure}
The MTF also depends only on one parameter being the number of bins which determines the resolution of the value grid. In this case, we chose 500 bins for the SAX algorithm, using the elbow method~\parencite{thorndike1953belongs} on the root-mean-square error of the inverse SAX transformation as a function of the number of used bins (as illustrated in Fig~\ref{fig:SAXbins}). It seems a good trade-off between capturing relevant patterns while excluding noise and computational burden.

\textbf{RP:}
The subsection length is fixed at 1 in order to avoid losing information and to achieve the required output dimensions.

\textbf{GS:}
For the GS encoding, we took the sub-series of size $512$ and turned them into $K=64$ series of size $K=64$, using a stride of $s=7$. This leads to $505$ values per sub-series in total. The remaining 7 points are discarded (3 at the beginning, 4 at the end). In addition to the original formulation, we explored three further variants: 1) $P=255$, 2) $P=1$, and 3) the min-max scaling only, that is, $P=1$  without rounding to the nearest integer. Since it turned out that the first and last variant were yielding extremely similar results, for reasons of clarity, only $P=1$ and the min-max variants are presented in the results. For all the three variants, we set the upper and lower bounds $\mathrm{UB}$ and $\mathrm{LB}$ to $1.2$, the maximum and the minimum observed over the training set, similarly to GAF. 

\textbf{SP:}
For the spectrogram, we propose to perform the STFT on the entire time series,  using a Hanning window of size 126 and a temporal stride of 8. The output is of size $64\times7\,680$ and is split into 120 slices of size $64\times64$. This approach is similar to that of first splitting the series before performing the STFT, but limits the border effects.

\textbf{SC:}
For the SC, the discrete wavelet transform with a Ricker wavelet was utilized. The 64 different scales of the window were obtained by the rule illustrated in equation \ref{eq:sc_scale}

\begin{equation}
             scale_j=2^{\frac{j}{4}};\quad j\in [1,64]
             \label{eq:sc_scale}
\end{equation}

The SC output of size $64\times512$ was then downscaled by factor of 8 along the horizontal axis using average pooling to achieve the required dimension of 64x64. Based on visual inspection, these parameters have been shown to capture the general time series patterns sufficiently well.
    
Fig.~\ref{fig:examples} illustrates, for the same time series, the different images obtained after each encoding.
    
    \subsection{Hyperparameters of the DL algorithms}
    \begin{table}
    \caption{Architecture of the two networks: 1D CAE and 2D CAE. The decoder always exactly mirrors the encoder.}\label{tab:1d_arch}
    \centering
        \begin{tabular}{l|lccc}
        \toprule
        & & \multicolumn{2}{c}{Kernel} & \\
        Section & Layer & Size & Stride & Shape \\ 
        \midrule 
        \multicolumn{5}{ c }{\textbf{1D-CAE Architecture}} \\
        \midrule
        \multirow{7}{*}{Encoder} 
         & Input & -  & -  & 1x512x1 \\
         & Conv1 & 16 & 4  & 1x128x64 \\
         & Pool1 &  2 & 2  & 1x64x64 \\ 
         & Conv2 &  8 & 2  & 1x32x128 \\
         & Pool2 &  2 & 2  & 1x16x128 \\ 
         & Conv3 &  4 & 2  & 1x8x256 \\
         & Pool3 &  2 & 2  & 1x4x256 \\ \midrule
        \multirow{3}{*}{\shortstack[l]{Fully-\\Connected}} 
         & Unfold     & - & -   & 1\,024 \\
         & Bottleneck & - & -   & 300\\
         & Dense2     & - & -   & 1\,024 \\ \midrule
        \multirow{1}{*}{Decoder}
          & \multicolumn{4}{ c }{Mirrors the architecture of the Encoder} \\
        \midrule
        \multicolumn{5}{ c }{\textbf{2D-CAE Architecture}} \\
        \midrule
        \multirow{3}{*}{Encoder} 
         & Input & - & -     & 64x64x1 \\
         & Conv1 & 2 & 2  & 32x32x64 \\
         & Conv2 & 2 & 2  & 16x16x128 \\ \midrule
        \multirow{3}{*}{\shortstack[l]{Fully-\\Connected}} 
         & Unfold     & - & -  & 32\,768\\
         & Bottleneck & -  & -  & 300\\
         & Dense2     & -  & - & 32\,768 \\ \midrule
        \multirow{1}{*}{Decoder}
          & \multicolumn{4}{ c }{Mirrors the architecture of the Encoder} \\
          \bottomrule
        \end{tabular}
    \end{table}   
    For both 1D and 2D auto-encoders, the chosen architectures are detailed in Table 1. They correspond to the one minimising the mean square error on the reconstruction loss, when performing 5-fold cross validation on the healthy set. The searched parameters consist in the number of layers, from 1 to 3, the bottleneck size, from 100 to 500 in steps of 100, the number of filters, from $2^2$ to $2^9$, and the kernel size (2, 4 or 8). A Leaky ReLU activation function is used with alpha equal to 0.3.
    We use the Adam optimizer with a learning rate of 0.001, $\beta_1$ equal to 0.9 and mean squared error as the loss function. Additionally, both networks are trained for 10 epochs with a batch size of 200. This accounts for 1\,677 sequences $\times$ 120 slices per sequence $\times$ 10 epochs or 2\,012\,400 iterations, a sufficient number to observe the network training loss convergence. The convergence of the network is confirmed by the low standard deviations of the results over 5 repetitions of the experiments, as illustrated in Table~\ref{tab:scores}.
    
    \subsection{Results}
We compare, in  Table~\ref{tab:scores}, the different models based on the average true and false positive rates (TPR, FPR), the F1 score, the area under the receiving operator curve (AUC) and the time required to encode 10\,000 images over 5 runs. We chose to report the TPR and FPR, in particular, since in the context of practical application, the FPR indicates whether the expected rate of false alarms is acceptable, while the TPR indicates the detection success of the algorithm. Also, from the number of samples and these two metrics, all other metrics found in the literature can be computed.
 
\begin{figure*}
\centering
\includegraphics[width=0.28\textwidth]{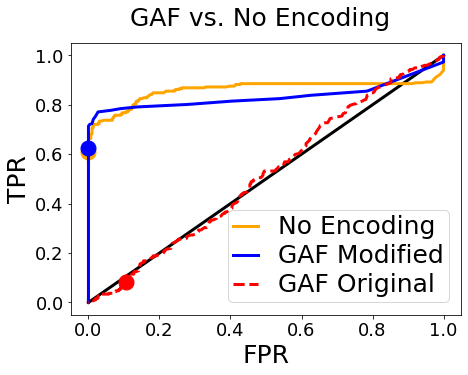}
\includegraphics[width=0.28\textwidth]{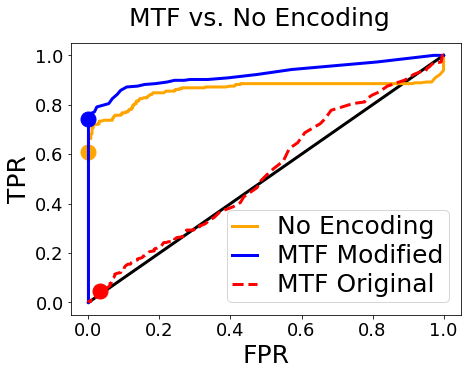}
\includegraphics[width=0.28\textwidth]{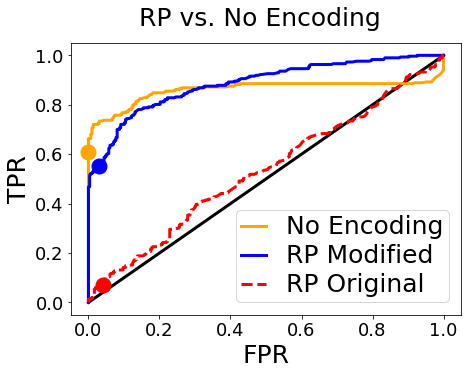}\\
\includegraphics[width=0.28\textwidth]{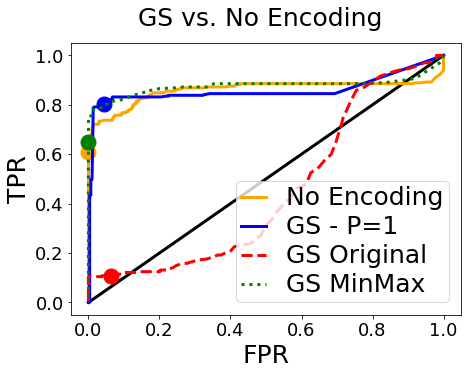}
\includegraphics[width=0.28\textwidth]{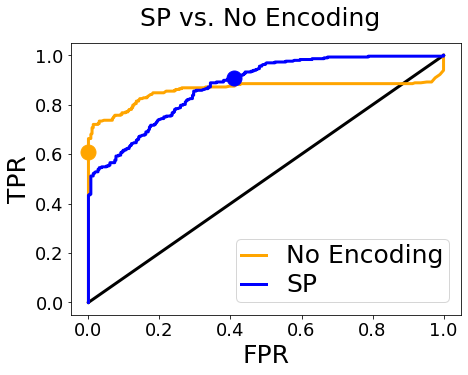}
\includegraphics[width=0.28\textwidth]{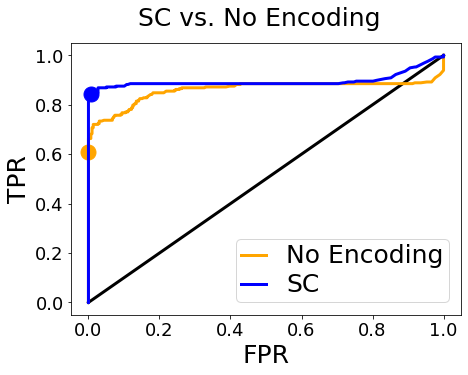}
\caption{\textbf{ROC curves}. This figure illustrates the ROC curves for the 2D model subjected to all individual encodings, as well as for the 1D model. The black diagonal line represents the random number generator baseline. 
\label{fig:roc}}
\end{figure*}

 The F1 score is defined as:
 \begin{equation}
     F1 = \frac{2 \cdot \mathrm{TP}}{2\cdot\mathrm{TP}+\mathrm{FP}+\mathrm{FN}},
 \end{equation}
The AUC is the area under the receiving operator curve (ROC) for the six encodings and the proposed extensions are shown in Figure~\ref{fig:roc}.

 \begin{table}
\centering
\caption{For the different models, originals (Or.) and modified (Mod.), the true and false positive rates (TPR and FPR), the F1 score, the area under the curve (AUC), and the time required to encode 10\,000 slices.
}
\label{tab:scores}
\resizebox{\ifdim\width>\columnwidth
        \columnwidth
      \else
        \width
      \fi}{!}{%
\begin{tabular}{ l|cccc|c}
     \toprule
     \multicolumn{6}{c}{\textbf{Performance Measures (\%)}}\\
     \midrule
      Encodings & TPR  & FPR  & F1  & AUC & T[s] \\
      \midrule
      No Enc. & 62 $\pm$ 01  & \textbf{01} $\pm$ 02 & 76 $\pm$ 01 & 81 $\pm$ 01 & -\\
     GAF Or.  & 12 $\pm$ 04 & 13 $\pm$ 03 & 20 $\pm$ 07 & 5 $\pm$ 01  & 24 \\
     GAF Mod.   & 68 $\pm$ 03 & 03 $\pm$ 00 & 81 $\pm$ 02 & 84 $\pm$ 02 & 27\\
     MTF Or.   & 04 $\pm$ 00 & 03 $\pm$ 00 & 08 $\pm$ 01 & 51 $\pm$ 00  & 34 \\
     MTF Mod.   & 73 $\pm$ 01 & \textbf{00} $\pm$ 00 & 85 $\pm$ 00 & 87 $\pm$ 00  & 264 \\
     RP Or.   & 07 $\pm$ 01 & 04 $\pm$ 01 & 13 $\pm$ 01 & 52 $\pm$ 00  & 34 \\
     RP Mod.   & 54 $\pm$ 04 & 03 $\pm$ 01 & 68 $\pm$ 03 & 75 $\pm$ 02  & 36 \\
     SP   & \textbf{84} $\pm$ 07 & 42 $\pm$ 03 & 74 $\pm$ 04 & 71 $\pm$ 05  & 4 \\
     SC   & \textbf{85} $\pm$ 02 & 01 $\pm$ 00 & \textbf{91} $\pm$ 01 & \textbf{92} $\pm$ 01 & 62 \\
     GS Or.   & 11 $\pm$ 00 & 06 $\pm$ 00 & 18 $\pm$ 00 & 52 $\pm$ 00  & 1 \\
     GS (P=1)   & 80 $\pm$ 00 & 04 $\pm$ 00 & 87 $\pm$ 00 & 89 $\pm$ 00  & 1 \\
     GS MM  & 63 $\pm$ 01 & \textbf{00} $\pm$ 00 & 77 $\pm$ 01 & 81 $\pm$ 01  & 1 \\
    \bottomrule
    \end{tabular}}
\end{table}
 
\begin{figure*}
\centering
\includegraphics[width=0.28\textwidth]{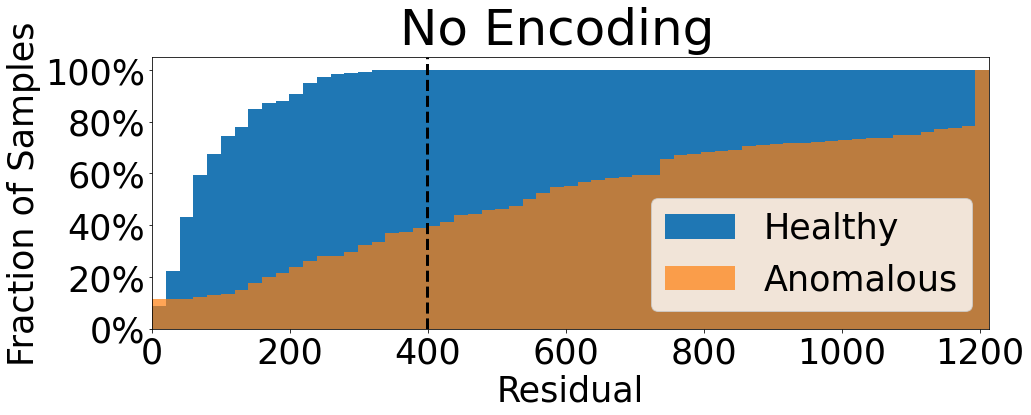}
\includegraphics[width=0.28\textwidth]{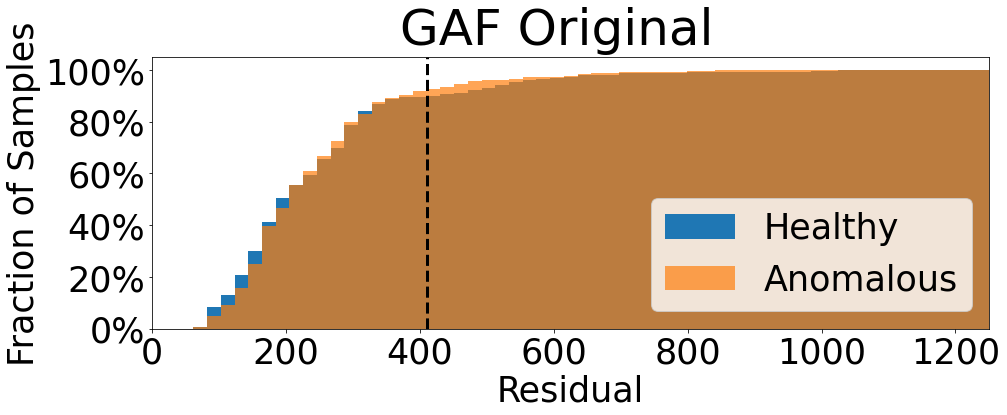}
\includegraphics[width=0.28\textwidth]{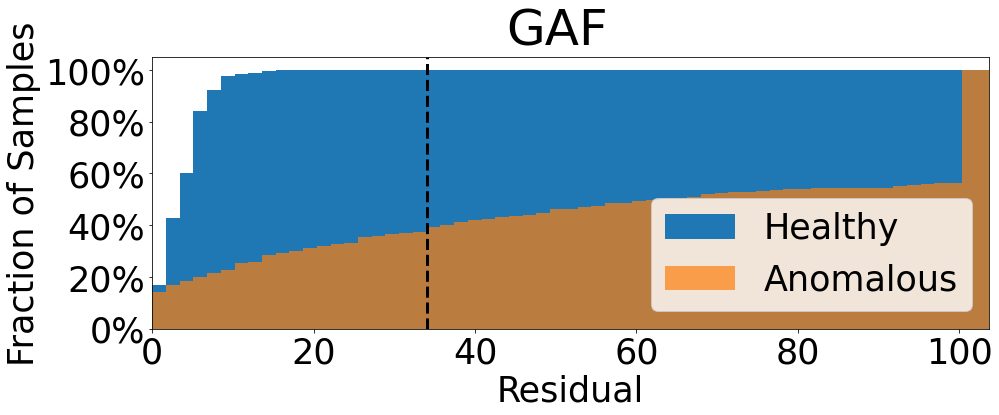}\\
\includegraphics[width=0.28\textwidth]{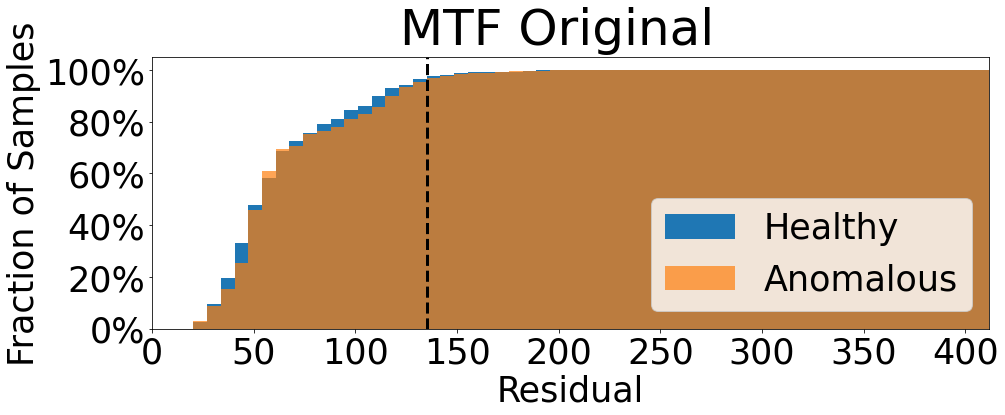}
\includegraphics[width=0.28\textwidth]{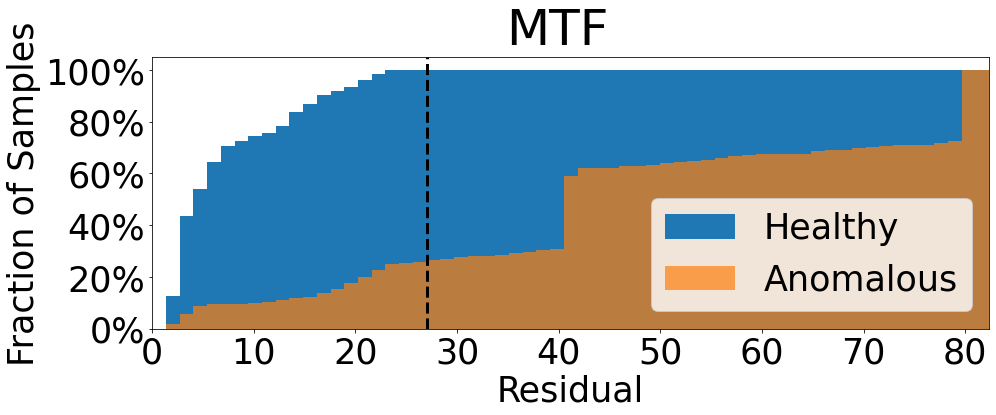}\\
\includegraphics[width=0.28\textwidth]{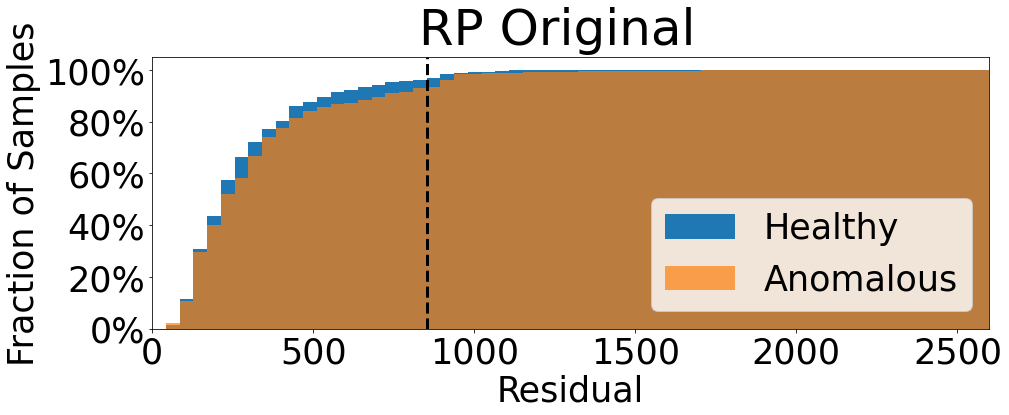}
\includegraphics[width=0.28\textwidth]{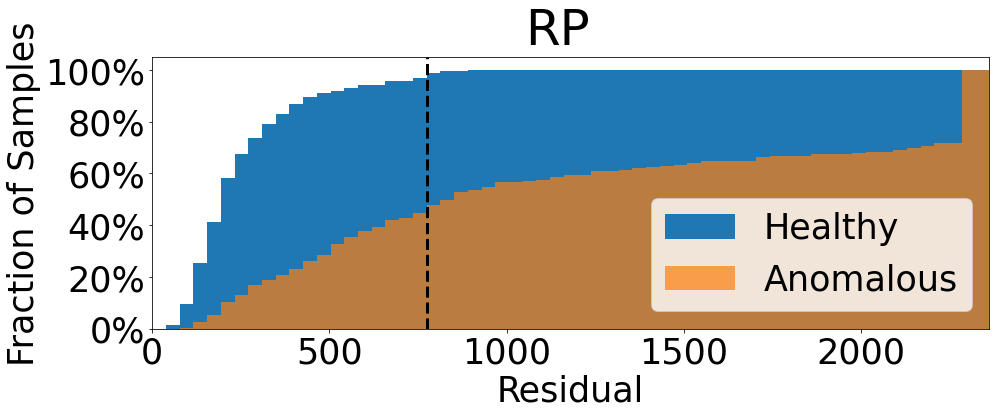}\\
\includegraphics[width=0.28\textwidth]{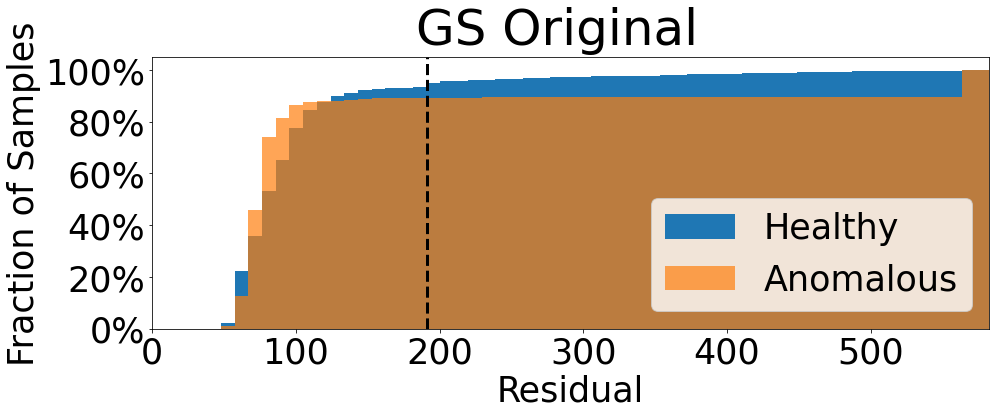}
\includegraphics[width=0.28\textwidth]{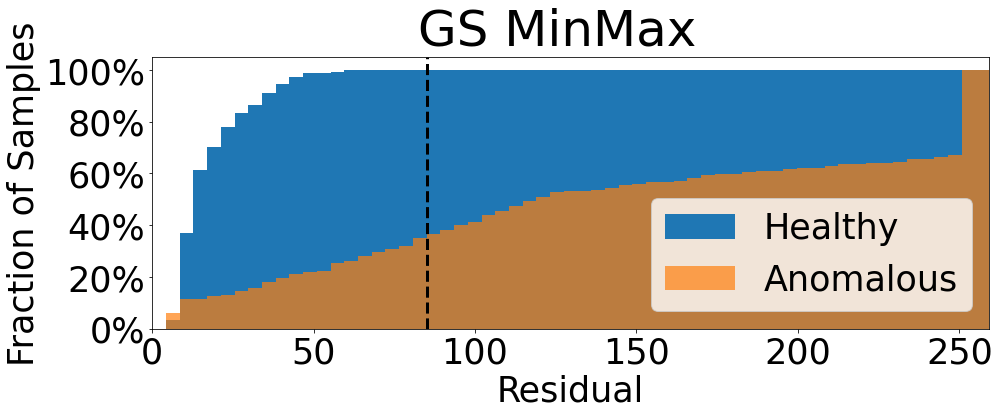}
\includegraphics[width=0.28\textwidth]{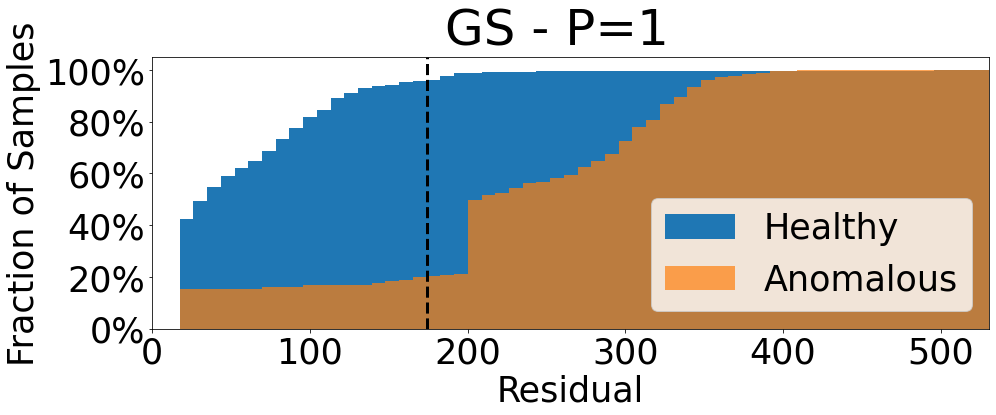}\\
\includegraphics[width=0.28\textwidth]{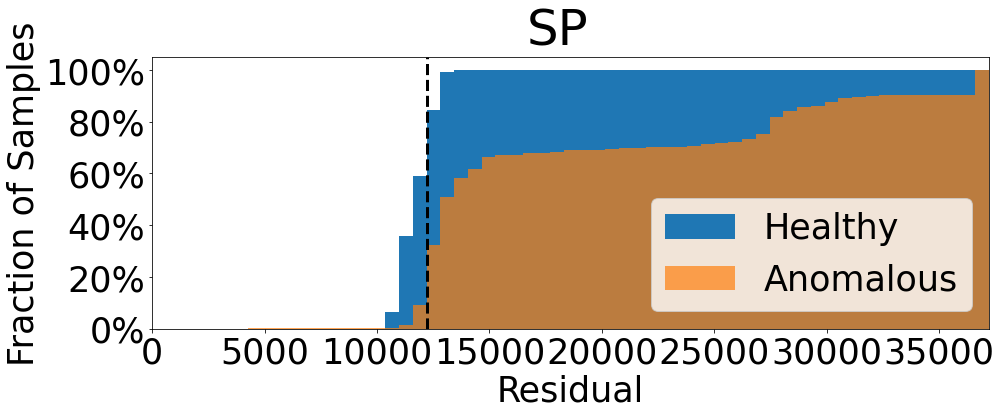}
\includegraphics[width=0.28\textwidth]{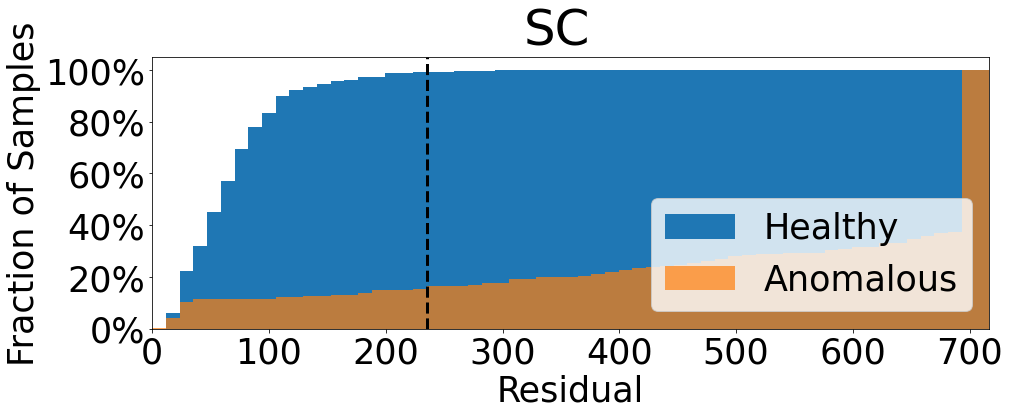}
\caption{\textbf{Residual Cumulative Distribution}. This figure illustrates the residual cumulative distribution for each encoding. The vertical dotted line represents the threshold. The last bin is for residuals beyond the boundary of the plots.
\label{fig:res}}
 \end{figure*}%

From the analysis of Figure \ref{fig:roc},  \ref{fig:res} and Table~\ref{tab:scores}, we can conclude that for the case study dataset and under the proposed setup, the scalogram is the best-performing approach. It achieves the highest scores for the F1 and AUC metrics. 

Furthermore, all proposed modifications to the encodings GAF, MTF, RP, and GS improve the performance of the anomaly detection algorithms as compared to their original formulations. Most modifications involved making the transformations dependent on the whole training set, instead of its sliced input only, as originally proposed. This allows us to better capture the characteristic behaviour of the main class, since the transformation is designed to take the whole dataset into account rather than single slices. For the GS encoding, finding the right $P$ value can have a significant impact on the results. Finding this parameter \textit{a priori} is a difficult task. Additionally, not using the conversion to integers also outperforms the baseline. The choice of these parameters is, therefore, sensible in a purely unsupervised task.

Based on the F1 metric comparison, all encodings in their modified versions outperform the baseline. Note that the F1 metric is computed based on the proposed threshold as defined above. The spectrogram and the recurrence plots have a lower AUC than the baseline, which means that with a different setting of the threshold, their advantage is not certain. However, setting this threshold optimally in an unsupervised detection task is an open research question. 

At the dataset level, all approaches achieve very low FPR, highlighting that the main class has been well learned. Since the networks were designed and trained based on healthy data only, this finding validates their ability to characterise the healthy conditions. The main differences between the different time series-to-image encodings are mostly based on their ability to better discriminate among the different types of anomalies.  The difference in performance may be due to the fact that some encodings are better suited for some types of anomalies. 

While the results show that time series-to-image encodings can provide benefits in terms of performance and partly also in terms of interpretability, at least for the spectrogram and scalogram whose visual analysis is common in the literature and is familiar to many domain experts, using them comes at the expense of increased computational complexity and required memory. Especially the SC and the MTF, which coincidentally are among the best-performing encodings, require a considerable amount of time to encode large numbers of time series. For the MTF, this shortcoming only holds for the bin assignment process, which is part of the original formulation. The proposed extension with the SAX algorithm overcomes this limitation. 
        
\section{Discussion}

The results presented above demonstrate the validity of the approach since the modified GAF, MTF, GS, and the scalogram offer an improvement on the TPR, the F1 score, and the AUC compared to the baseline. The results also open up interesting points for discussion and perspectives for further development and evaluation.

\textbf{Defining the transformation:} Using the original formulations of the different encodings~\parencite{JMLRpyts} in our case study led to poor performance, as reflected in Table~\ref{tab:scores}. Defining the transformation parameters per input tends to make the encoded images more uniform. While it might be beneficial in a supervised setup to learn the very specific class discriminative features and differences in the images, it is prejudicial in an unsupervised setup, based on the learning of the characteristics of a single (healthy) class. Taking the whole dataset into account when performing the transformation was an essential step in the performance of the proposed encodings.

\textbf{Threshold choice:} Learning the right threshold using only healthy data is always a difficult task. In our work, the choice of the 99th percentile on the residuals seems to yield reasonable results, which are in line with the threshold-independent metric AUC. However, we highlight the fact that this threshold has been learnt on normal data only. Therefore, it is difficult to state with confidence that this choice will be effective for any other problem, any other type of data and anomalies. In future research, using these approaches in an ensemble learning approach may provide additional insights and improve the performance.

\textbf{Selection of time-series to image encoding:} Even though all image encodings provide reasonably good results, it is still not easy to say beforehand which encoding will be the best for a particular use case. It is quite obvious that each encoding will be more adapted to certain types of time series and will perform differently on various types of anomalies. 

\textbf{Architecture of the CAE:} We used a rather simple convolutional neural network since the main goal was a comparison between the different encodings and the same architecture was applied to all the encodings. It could be interesting to explore other architectures to study their impact on the performance. 

\textbf{Aggregation of residuals:} The computation of the encodings can be quite costly when the number of time steps increases. Computational performance on the same number of time series with the same hardware is shown in Table \ref{tab:scores}. This scalability of the image encodings is one of the reasons why we cut the initial time series into smaller ones. Since the other reason was the ability to detect local anomalies, we used here the maximum residuals on all slices and compared them to the threshold. Other choices were available (mean, quantile over the slices, etc.) and would have impacted the results. Further exploring the choice of the decision criteria could be a subject for future investigations. Since the main goal of this study was to compare the different encodings, the same decision was applied to all encodings. 

\textbf{Interpretability of image encodings.} Since some of the image encodings, used in a deep learning approach, yielded competitive results in the anomaly detection task, it means that the corresponding images contain the information necessary to discriminate between healthy and anomalous signals. In an application environment, domain experts may want to verify or manually further analyse the automatically classified patterns. In the case of manual analysis, analysing the images may prove to be more intuitive compared to pure time series data in terms of identifying, interpreting, and tracking patterns and anomalies on different time scales. Some domains already use images to analyse their signals, in particular spectrograms and scalograms. Typically, domain experts develop an intuition and experience in identifying different types of anomalies in such representations. Using other encodings could also become quite natural if the experts are willing to train themselves to recognize different conditions from the generated images, although the image relevance will depend on the application and would require further research.

\section{Conclusion and Perspectives}

In this paper, we evaluated the application of several time series-to-image encodings in an unsupervised anomaly detection setting. In addition to applying the originally proposed encodings, we also suggested modifications to the GAF, MTF, and GS encodings to make them more suitable within the proposed framework for anomaly detection. 
Testing all approaches, including the use of the raw time series directly, in similar conditions and on the exact same data, we demonstrated that all the (modified) image encodings led to an improvement as compared to using the raw time series. 
The good results achieved by some of the encodings might be related to the fact that converting the time series into two-dimensional representations allows one to capture more complex patterns between measurements, such as correlations, recursive behaviors, or spectral components. These relationships can then improve the ability to characterise healthy data and to separate anomalies. In comparing the encodings and their modified versions on the same benchmark, and in an unsupervised setup, these results complement the previous studies conducted in supervised setups that already hinted the benefits of these approaches. 

In the present paper, the framework was tested on a specific type of data, i.e., vibration data from in-flight measurements. Future research is required to evaluate the benefits in other contexts, with other types of data, and with other anomalies. Ideally, an analysis exploring which encoding would best suit which type of data and which type of anomaly would be extremely beneficial to the community. Based on such analysis, the encodings could be combined in an ensemble learning framework to perform both detection and diagnostics.

\printbibliography

\end{document}